\documentclass[10pt,journal,compsoc]{IEEEtran}
\usepackage{comment}
\usepackage{wrapfig}
\usepackage{subfigure}
\usepackage{graphicx}
\usepackage{hyperref}
\hypersetup{
    colorlinks=true,
    linkcolor=blue,
    filecolor=magenta,      
    urlcolor=cyan,
}

\usepackage{draftwatermark}
\SetWatermarkText{Preprint, Accepted by IEEE Journal on Emerging and Selected Topics in Circuits and Systems}
\SetWatermarkScale{0.1}

\begin{document}

\title{Low-Power Computer Vision: Status, Challenges, Opportunities}

\author{
Sergei Alyamkin,
 Matthew Ardi,
 Alexander C. Berg,
 Achille Brighton,
 Bo Chen,
 Yiran Chen,
 Hsin-Pai Cheng,
 Zichen Fan,
 Chen Feng,
 Bo Fu,
 Kent Gauen,
 Abhinav Goel,
 Alexander Goncharenko,
 Xuyang Guo,
 Soonhoi Ha,
 Andrew Howard,
 Xiao Hu,
 Yuanjun Huang,
 Donghyun Kang,
 Jaeyoun Kim,
 Jong Gook Ko,
 Alexander Kondratyev,
 Junhyeok Lee,
 Seungjae Lee,
 Suwoong Lee,
 Zichao Li,
 Zhiyu Liang,
 Juzheng Liu,
 Xin Liu,
 Yang~Lu,
 Yung-Hsiang Lu\thanks{Corresponding Author: Yung-Hsiang Lu, Email: yunglu@purdue.edu},
 Deeptanshu Malik,
 Hong Hanh Nguyen,
 Eunbyung Park,
 Denis Repin,
 Liang Shen,
 Tao Sheng,
 Fei~Sun,
 David~Svitov,
 George K. Thiruvathukal,
 Baiwu Zhang,
 Jingchi Zhang,
 Xiaopeng Zhang,
 Shaojie Zhuo
}

\markboth{Journal on Emerging and Selected Topics in Circuits and Systems}%
{Author \MakeLowercase{\textit{et al.}}}

\maketitle

\begin{abstract}
Computer vision has achieved impressive progress in recent years. Meanwhile, mobile phones have become the primary computing platforms for millions of people. In addition to mobile phones, many autonomous systems rely on visual data for making decisions and some of these systems have limited energy (such as unmanned aerial vehicles also called drones and mobile robots). These systems rely on batteries and energy efficiency is critical. This article serves two main purposes: (1) Examine the state-of-the-art for low-power solutions to detect objects in images. Since 2015, the IEEE Annual International Low-Power Image Recognition Challenge (LPIRC) has been held to identify the most energy-efficient computer vision solutions. This article summarizes 2018 winners' solutions. (2) Suggest directions for research as well as  opportunities for low-power  computer vision. 
\end{abstract}

\begin{IEEEkeywords}
Computer Vision, Low-Power Electronics, Object Detection, Machine Intelligence
\end{IEEEkeywords}

\IEEEpeerreviewmaketitle

\section{Introduction}

Competitions are an effective way of promoting innovation and system integration. “Grand Challenges” can push the boundaries of technologies and open new directions for research and development. The DARPA Grand Challenge in 2004 opened the era of autonomous vehicles. Ansari X Prize started a new era of space tourism. Since 2010, the ImageNet Large Scale Visual Recognition Challenge (ILSVRC) has become a popular benchmark in computer vision for detecting objects in images.  ILSVRC is an online competition: contestants submit their solutions through a website. The only factor for comparison is the accuracy  and there is no strict time limit. 

Phones with cameras appeared as early as year 2000.   Since 2011,  smartphones outsell personal computers and become the primary computing platforms for millions of people.   Computer vision technologies have become widely used on smartphones. For example, people use smartphones for comparison shopping by taking photographs of interested products and search for reviews, similar products, and prices. Face detection has become a standard feature on digital cameras.
Better computer vision technologies, however, are not the ``most desired'' features for future smartphones. Instead, longer battery life has consistently ranked as one of the most desired features. Recognizing the need for energy-efficient computer vision technologies, the IEEE Rebooting Computing Initiative started the first IEEE Low-Power Image Recognition Challenge (LPIRC) in 2015. LPIRC aims to identify energy-efficient solutions for computer vision. These solutions have a wide range of applications in mobile phones, drones, autonomous robots, or any intelligent systems equipped with digital cameras carrying limited energy. 

LPIRC is an annual competition identifying the best system-level solutions for detecting objects in images while using as little energy as possible~\cite{7858303, 7372672, 8342099, AIM2018, Lu2019}. Although many competitions are held every year, LPIRC is the only one integrating both image recognition and low power. In LPIRC, a contestant's system is connected to the referee system through an intranet (wired or wireless). There is no restriction on software or hardware. The contestant's system issues HTTP GET commands to retrieve images from the referee system and issues HTTP POST commands to send the answers. To reduce the overhead of network communication, the referee system allows retrieving 100 images at once as a zip file. Also, one HTTP POST command can send multiple detected objects from many images.
Each solution has 10 minutes to process all images (5,000 images in 2015 and 20,000 images since 2016). The scoring function is the ratio of accuracy (measured by {\it mean average precision}, mAP) and the energy consumption (measured by Watt-Hour).
Table~\ref{table:scores2015-2018} shows the champions' scores. As can be seen in this table, the scores improve 24 times from 2015 to 2018. The best mAP in LPIRC
is lower than the best score in the ImageNet Challenge (ILSVRC). The winner of the ImageNet Challenge (in 2017) achieved mAP of 0.731. The ImageNet Challenge has no time restriction, nor is energy considered. 

\begin{table}[h]
\centering
\begin{tabular}{|llllr|} \hline
Year & Accuracy & Energy & Score  & Ratio  \\ \hline
2015 & 0.02971  & 1.634  & 0.0182 & 1.0    \\
2016 & 0.03469  & 0.789  & 0.0440 & 2.4    \\
2017 & 0.24838  & 2.082  & 0.1193 & 6.6    \\
2018 (Track 2) & 0.38981  & 1.540  & 0.2531 & 13.9   \\
2018 (Track 3) & 0.18318  & 0.412  & 0.4446 & 24.5   \\ \hline 
\multicolumn{5}{l}{Accuracy: mean average precision, mAP} \\
\multicolumn{5}{l}{Energy: Watt-Hour} \\
\multicolumn{5}{l}{Score = Accuracy / Energy} \\
\multicolumn{5}{l}{Ratio: 2015 is the basis} \\
\end{tabular}
\caption{Scores of LPIRC champions since 2015.}
\label{table:scores2015-2018}
\end{table}

This paper has two major contributions: (1) It describes an international competition  in low-power image recognition and the winners have demonstrated significant progress. The importance of this competition is reflected by
the list of sponsors as well as the profiles of the teams. (2) The paper provides an overview of different methods that won the competition.  The winners explain the design decisions and the effects of these factors. 
This paper extends an earlier online version that focuses on the 2018 LPIRC~\cite{LPIRC2018arXiv} by adding the discussion of future research directions for low-power computer vision.  This paper is organized as follows. Section 2 describes the different tracks in LPIRC. Section 3 summarizes the scores. Section 4 explains the 2018 winners' solutions. Section 5  describes industry support for low-power computer vision.  Section 6 discusses  challenges and opportunities for low-power computer vision. Section 7 concludes the paper.

\section{Tracks of LPIRC}

In the past four years, LPIRC has experimented with different tracks with different rules and restrictions. In 2015, offloading of processing was allowed but only one team participated. In 2016, a track displayed images on a computer screen and a constantant's system used a camera (instead of using the network) to capture the images but only one team participated in this track. These two tracks are no longer offered.  
Since 2015, LPIRC has been an onsite competition: contestants  bring their systems to conferences (Design Automation Conference for 2015 and 2016, International  Conference on Computer Vision and Pattern Recognition for 2017 or 2018). Onsite competitions are necessary since LPIRC has no restrictions on hardware or software. Even though this gives contestants the most flexibility, the need to travel to the conferences with hardware potentially discourages some participants.  In 2018, two new tracks were created to encourage participation. These two tracks allowed online submissions and contestants did not have to travel.  The third track for 2018 LPIRC was the original track always been offered: no restriction on hardware or software.  The following sections describe the tracks in 2018 LPIRC. 

\subsection{Track 1: TfLite Model on Mobile Phones}

\begin{figure}[h]
\centering
\includegraphics[width=3in]{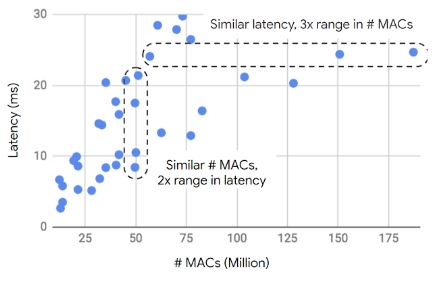}
\caption{Relationship between MACs and latency~\cite{ChenGilbert20180420}.
}
\label{fig:maclatency}
\end{figure}

This new track, also known as the On-device Visual Intelligence Competition, focused on model architecture: contestants submitted their inference neural network models  in TfLite format using Tensorflow.
The models were benchmarked on a fixed inference engine (TfLite) and hardware model (Pixel 2 XL phone). The task is ImageNet classification. The submissions should strive to classify correctly as many images as possible given a time budget of 30 ms per image. The submissions were evaluated on a single core with a batch-size of one to mimic realistic use cases on mobile devices.
Pixel 2 (and 2XL) are mobile phones running Android 8.0. Both use Qualcomm Snapdragon 835 processors and have 4GB memory.

\begin{figure}[h]
\centering
\includegraphics[width=3in]{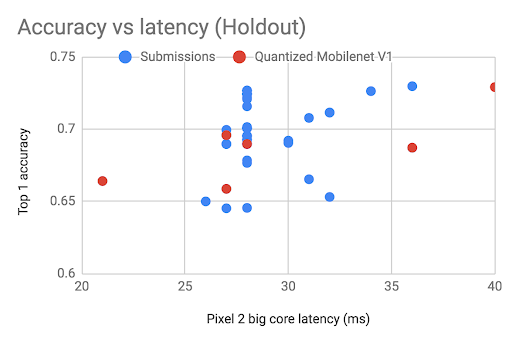}
\caption{Latency and scores in 2018 Track 1 for the holdout dataset.
}
\label{fig:track1holdout}
\end{figure}

This track provides a benchmarking platform  for repeatable measurements and fair comparison of mobile model architectures. The flexibility of submitting just the model and not the full system allows this track to be an online competition. This convenience helps to boost the submission count to 128 models within just 2 weeks. Although the scoring metric is based on inference latency rather than energy consumption, the two are usually correlated when the same benchmarks are used on the same hardware. Track 1's latency-based metric is critical to accelerating the development of mobile model architectures. Prior to this metric, there was no common, relevant, and verifiable metric to evaluate the inference speed of mobile model architectures for vision tasks. Many papers characterize inference efficiency using unspecified benchmarking environments or latency on desktop as a proxy. Even the most commonly used metric, MACs (multiply-add count) does not correlate well with inference latency in the real-time (under 30 ms) range~\cite{ChenGilbert20180420}, as shown in Figure~\ref{fig:maclatency}.
The high participation of this track established the new state-of-the-art in mobile model architecture design. Within the 30 ms latency range, the best of track 1 submissions outperformed the previous state-of-the-art based on quantized MobileNet V1, by 3.1\% on the holdout test set,
as shown in Figure~\ref{fig:track1holdout}.

The submissions have demonstrated a considerable amount of diversity. Slightly over half (51.7\%) of the solutions are quantized. Most of the  architectures (74.1\%) are variations of the existing Mobilenet model family, namely quantized V2 (22.4\%), quantized V1 (24.1\%) and float V2 (27.6\%). The predominant dependence on Mobilenets is not surprising, considering their exceptional performance on-device and the convenient support by TfLite. To sustain future research on mobile computer vision, Track 1 should also reward previously under- or un-explored model architectures. In future  LPIRC, Track 1 may  consider methods for facilitating the discovery of novel architectures and making the platform more inclusive to general vision researchers. 

\subsection{Track 2: Caffe 2 and TX 2}

To make online submission possible, Track 2 uses  Nvidia TX2 as the hardware platform and Caffe2 as the software platform so that the organizers may execute contestants' solutions and calculate  the scores.  To further assist contestants, a software development kit (SDK) 
is available~\cite{IEEELowPowera}. TX2 has NVIDIA Pascal GPU with 256 CUDA cores. The processor is HMP Dual Denver 2/2 MB L2 and Quad ARM® A57/2 MB L2. It has 8GB 128bit LPDDR4 memory with 59.7 GB/s data rate.  Caffe2 is a framework for deep learning supporting  GPUs and cloud through the cross-platform libraries. The other parts of Track 2 are identical to Track 3.

\subsection{Track 3: Onsite, No Restriction}

\begin{figure}[t]
\centering
\includegraphics[width=3in]{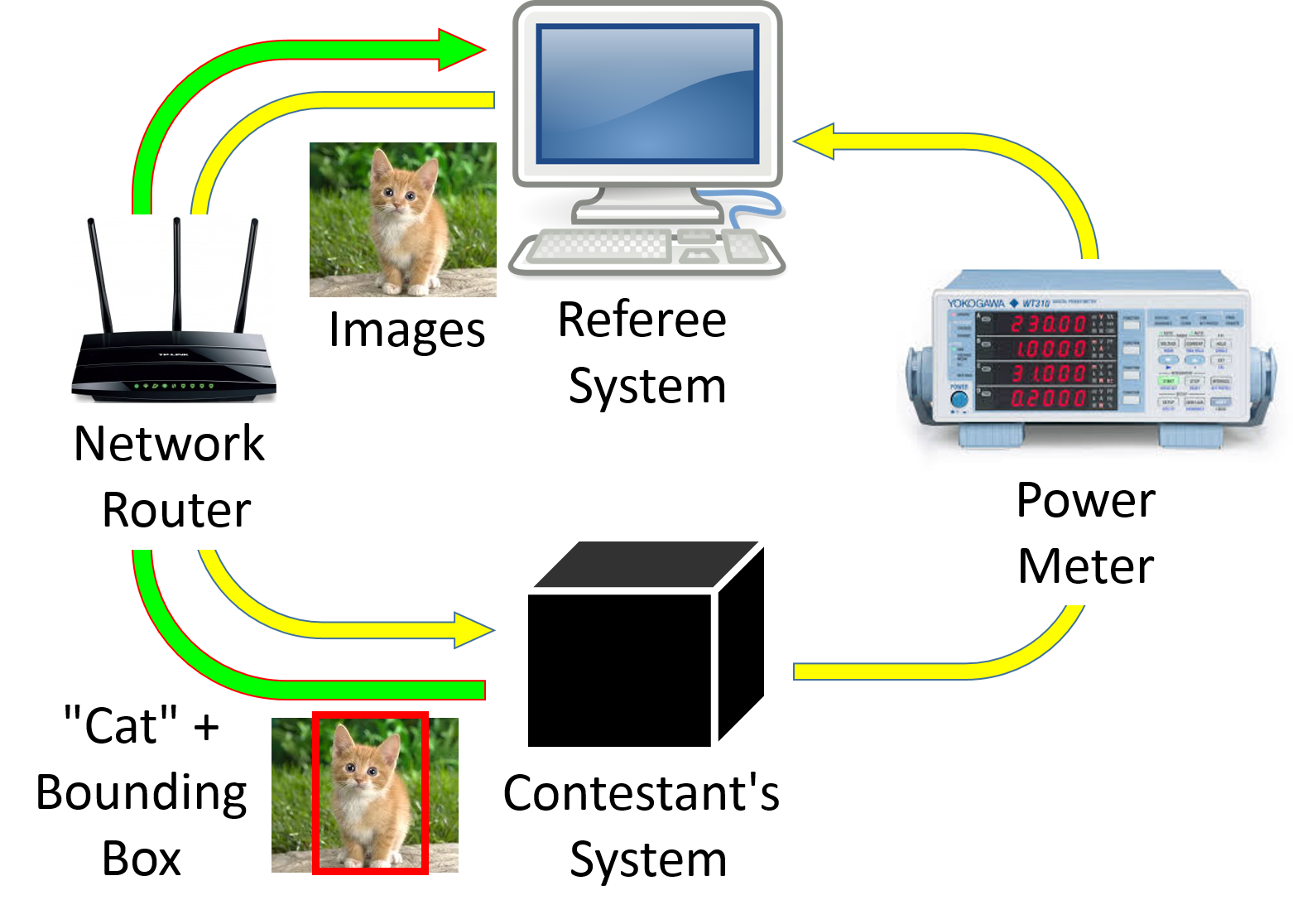}
\caption{Interactions between a contestant's system (the black box) and the referee system.
}\label{fig:track3}
\end{figure}

This is the original track always offered since 2015. This track has no restriction in hardware or software and gives contestants the most flexibility. Figure~\ref{fig:track3} illustrates the interactions between a contestant's system and the referee system.  The two systems are connected through a network router; both wireless and wired networks are available.  A system's energy consumption is measured using a high-speed power meter, Yokogawa WT310 Digital Power Meter. It can measure AC or DC and can synchronize with the referee system. Each team has 10 minutes to process all the images. There are  200 classes of objects in the competition (same as ImageNet). The training set is not restricted to ImageNet and contestants can use any datasets for training.
Test images are stored in the referee system and retrieved through the router. The results are uploaded to the referee system through the same network. After the contestant's system logs in, the power meter starts measuring the power consumption. The power meter stops after 10 minutes or when a contestant's system logs out. This allows a team that finishes all images within 10 minutes to reduce their  energy consumption. Additional information about Track 3 can be obtained from prior publications~\cite{7858303, 7372672, 8342099, AIM2018}. 

To win Track 2 or 3, a solution must be able to detect objects in images and mark the objects' locations in the images by {\it bounding boxes}. A successful object detection must identify the category correctly and the bounding box must overlap with the correct solution (also called ``ground truth'', marked by the LPIRC organizers) by at least 50\%.  A superior solution must be able to detect objects and  their locations in the images as fast as possible, while consuming as little energy as possible.

\subsection{Training  and Testing Data}
LPIRC uses the same training images as ILSVRC. For Track~1, the data from localization and classification task (LOC-CLS) is used. It consists of around 1.2 million  photographs, collected from Flickr and other search engines, hand-labeled with the presence or absence of 1,000 object categories. Bounding boxes for the objects are available, although Track 1 only considers classification accuracy as a performance metric. For tracks 2 and 3, the data from object detection is used. It has around 550,000 images and bounding boxes of labels for all 200 categories.

\begin{figure}[h]
\centering
\includegraphics[width=3in]{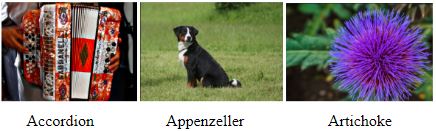}
\includegraphics[width=3in]{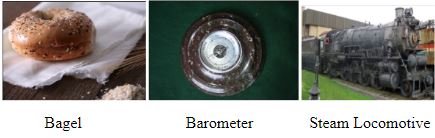}
\caption{Sample test images for Track 1.
}\label{fig:track1test}
\end{figure}

For testing data, Track 1 used newly created data with the ILSVRC image crawling tool to collect 100 images for each category from Flickr. When crawling the images, synonyms are used so that   relevant images are classified into the corresponding categories. The competition uses only the images uploaded after June 2017. Thumbnail images are used to remove duplicates by resizing the images to 30 $\times$ 30, and calculating the L2-norm of differences with images in the previous dataset. Ten representative images are manually chosen from each category. Representative images refer to the images that can be identified as one of the 1000 ImageNet classification categories without ambiguity. Figure~\ref{fig:track1test} shows samples test images for Track 1.
Tracks 2 and 3 also use the crawling tool to obtain images from Flickr with the context query phrases (e.g., ``cat in the kitchen'') instead of the label names (e.g., cat) as the keywords. The tool retrieves images with various types of objects. The annotations (bounding boxes) are created by using Amazon Mechanical Turk. 

\subsection{Other Datasets for Image Recognition}

For reference, this section describes some popular datasets used in computer vision:  PASCAL VOC, COCO (Common Objects in COntext), SUN (Scene UNderstanding), INRIA Pedestrian Dataset, KITTI Vision Benchmark Suite, and Caltech Pedestrian Dataset. Over time the image datasets have improved in two primary ways. First, the quality of the image annotations has significantly improved due to more sophisticated methods for crowdsourcing. Second, the variety of the dataset has increased, in both content and annotations, by increasing the number of classes.  

\begin{figure}[h]
\centering
\subfigure[]
{\includegraphics[height=1.1in]{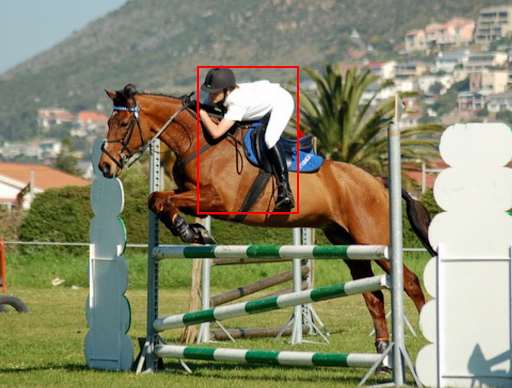}}
\subfigure[]
{\includegraphics[height=1.1in]{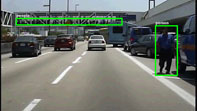}}
\subfigure[]
{\includegraphics[height=0.95in]{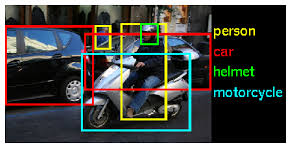}}
\subfigure[]
{\includegraphics[height=0.95in]{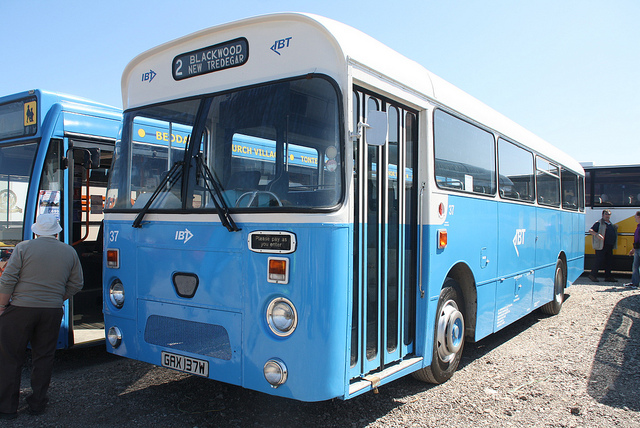}}
\caption{Sample images (a) PSACAL VOC,  (b) Caltech, (c) ImageNet, (d) COCO.
}\label{fig:sampleimages}
\end{figure}

PASCAL VOC started its first challenge in 2005 for object detection and classification of four classes. In 2012, the final year of the competition, the PASCAL VOC training and validation datasets consisted of 27,450  objects in 11,530 images with 20 classes. From 2009 - 2005 the overall classification average precision improved from 0.65 - 0.82 and detection average precision improved from 0.28 - 0.41~\cite{Everingham:2015:PVO:2725268.2725369}.
The COCO competition continues to be held annually with a primary focus of correctness. COCO contains over 2.5 million labeled instances in over 382,000 images with 80 common objects for instance segmentation~\cite{10.1007/978-3-319-10602-1_48}. 
The performance of a model on COCO has improved for bounding-box object detection from 0.373 to 0.525 during 2015 to 2017. 

\section{2018 LPIRC Scores}

\subsection{Track 1 Scores}

Table~\ref{table:track1winner}  shows the score of the winner of Track 1. It uses ImageNet for validation. The holdout set is freshly collected for the purpose of this competition.  The terms are defined as

\begin{itemize}
\item Latency (ms):  single-threaded, non-batch runtime measured on a single Pixel 2 big core of classifying one image

\item Test metric (main metric): total number of images corrected in a wall time of 30 ms divided by the total number of test images

\item Accuracy on Classified: accuracy in [0, 1] computed based only on the images classified within the wall-time

\item  Accuracy/Time: ratio of the accuracy and either the total inference time or the wall-time, whichever is longer

\item \# Classified: number of images classified within the wall-time

\end{itemize}

\begin{table}[h]
\centering
\begin{tabular}{|l|r|r|} \hline

& Validation Set & Holdout Set \\ \hline
Latency & 28.0 & 27.0 \\
Test Metric & 0.64705 & 0.72673 \\
Accuracy on Classified & 0.64705 & 0.72673 \\
Accuracy / Time & 1.08 E-06 & 2.22 E-06 \\
\# Classified & 20000 & 10927 \\ \hline
\end{tabular}
\vspace{0.1in}
\caption{Track 1 Winner's Score}
\label{table:track1winner}
\end{table}

\begin{figure}[h]
\centering
{\includegraphics[width=3in]{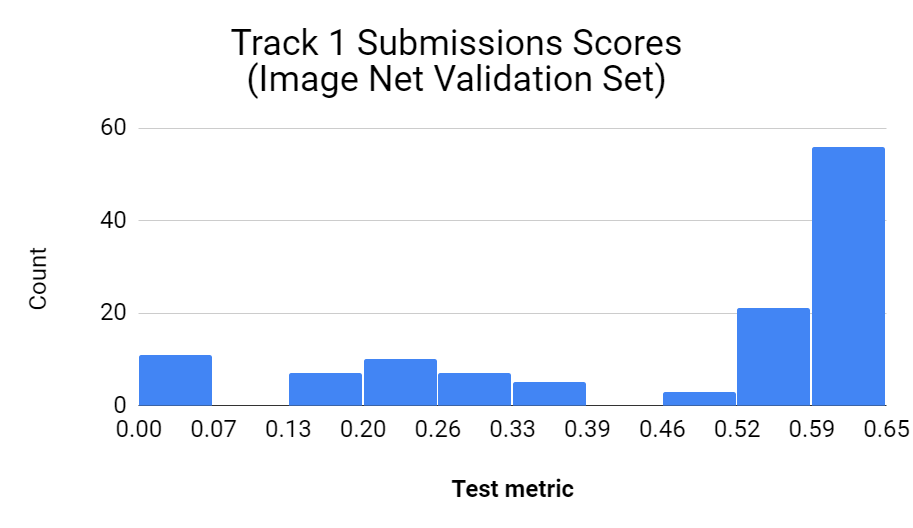}}
\caption{Track 1 scores. }\label{fig:track1scores}
\end{figure}

Track 1 received 128 submissions. Figure~\ref{fig:track1scores} represents a total of 121 valid submissions (submissions that passed the bazel test and successfully evaluated). Among them, 56 submissions received test metric scores between 0.59 and 0.65. The mean is 
0.4610; the median is
0.5654; the mode is
0.6347; the standard deviation
is 0.2119.

Some duplicate submissions that are successfully evaluated have differences with the “Test Metrics” score due to changes in the evaluation server. These changes are made throughout the progress of the competition. Files that have the same {\tt md5} hash information are considered duplicates. Duplicates may be caused by
submissions by the same team using different accounts. 
After eliminating duplicates, there are 97 unique submissions.

\begin{table}    \centering
    \begin{tabular}{|l|r|r|r|} \hline
Prize & mAP & Energy & Score \\ \hline
Winner & 0.3898 & 1.540 & 0.2531 \\
Second & 0.1646 & 2.6697 & 0.0616 \\
Third & 0.0315 & 1.2348 & 0.0255 \\ \hline
    \end{tabular}
    \vspace{0.1in}
    \caption{Track 2 Winners}
    \label{table:track2winners}
\end{table}

\begin{table}    \centering
    \begin{tabular}{|l|r|r|r|} \hline
Prize & mAP & Energy & Score \\ \hline
Winner & 0.1832 & 0.4120 & 0.44462 \\
Second & 0.2119 & 0.5338 & 0.39701 \\
Second &  0.3753 & 0.9463 & 0.39664 \\
Third & 0.2235 & 1.5355 & 0.14556 \\ \hline
    \end{tabular}
    \vspace{0.1in}
    \caption{Track 3 Winners}
    \label{table:track3winners}
\end{table}

\subsection{Tracks 2 and 3 Scores}

Table~\ref{table:track2winners} shows the scores of Track 2's winners. The champion achieved very high accuracy and finished recognizing all 20,000 images in less than 6 minutes. The 2018 winner's score is twice as high as the 2017 best score. Table~\ref{table:track3winners} shows the scores for Track 3's winners. Two teams' scores are very close and both teams win the second prize. Also, the top three scores outperform that of the 2017 champion. Two teams' energy consumption is much lower than previous years' champion's.  It should be noted that the champion in Track 3 has lower mAP with much lower energy consumption, thus a higher score.

\section{2018 Winners' Solutions}

This section describes the winners' solutions. Some winners decided not to share their solutions in this paper.

\subsection{First Prize of Track 1}

The Qualcomm team wins the first prize of Track 1. 
Qualcomm provides edge AI platforms with Snapdragon (including the Google Pixel 2 phone used in this competition).
Accurate and fast image recognition on edge devices requires several steps. First, a neural network model needs to be built and trained to identify and classify images 
Then, the model should run as accurate  and fast as possible.
Most neural networks are trained on floating-point models and usually need to be converted to fixed-point  to efficiently run on edge devices.

\begin{table*}[h]
\centering
    \begin{tabular}{|p{0.6in}|p{0.5in}|p{0.3in}|p{0.4in}|p{0.5in}|p{0.5in}|} \hline
         Neural network architecture &
Input image resolution &
Data type &
Accuracy (\%) &
Google Pixel-2 inference time (ms) & 
Accuracy/ inference time (\%/ms) \\ \hline

mobilenet v1 &
224x224 & 
float32 &
70.2 &
81.5 &
0.86 \\

mobilenet v1 &
224x224 &
uint8 &
65.5 &
68.0 & 
0.96 \\

mobilenet v1 & 
128x128 & 
uint8 & 
64.1 & 
28.0 & 
2.28 \\

mobilenet v2 & 
150x150 & 
uint8 & 
64.4 &
36.6 &
1.75 \\

mobilenet v2 &
132x132 &
uint8 &
62.7 &
31.8 &
1.97 \\

mobilenet v2 &
130x130 &
uint8 &
59.9 &
31.2 &
1.91 \\ \hline

    \end{tabular}
    \vspace{0.1in}
    \caption{Different configurations and results by the Expasoft team.}
    \label{table:track1thirdprize}
\end{table*}

For this competition, the model is based on MobileNet V2, but is modified to be quantization-friendly. Although Google's MobileNet models successfully reduce parameter sizes and computation latency due to the use of separable convolution, directly quantizing a pre-trained MobileNet v2 model can cause large precision loss. The team analyzes and identifies the root cause of accuracy loss due to quantization in such separable convolution networks, and solves it properly without utilizing quantization-aware re-training.
In separable convolutions, depthwise convolution is applied on each channel independently. However, the min and max values used for weights quantization are taken collectively from all channels. An outlier in one channel may cause a huge quantization loss for the whole model due to an enlarged data range. Without correlation crossing channels, depthwise convolution may be prone to produce all-zero values of weights in one channel. This is commonly observed in both MobileNet v1 and v2 models. All-zero values in one channel means small variance. A large “scale” value for that specific channel would be expected while applying batch normalization transform directly after depthwise convolution. This hurts the representation power of the whole model.

As a solution, the team proposes an effective quantization-friendly separable convolution architecture, where the nonlinear operations (both batch normalization and ReLU6) between depthwise and pointwise convolution layers are all removed, letting the network learn proper weights to handle the batch normalization transform directly. In addition, ReLU6 is replaced with ReLU in all pointwise convolution layers. From various experiments in MobileNet v1 and v2 models, this architecture shows a significant accuracy boost in the 8-bit quantized pipeline.\\

Using fixed-point inferencing while preserving a high level of accuracy is the key to enable deep learning use cases on low power edge devices. The team identifies the industry-wide quantization issue, analyzes the root cause, and solves it on MobielNets efficiently. The quantized modified MobileNet\_v2\_1.0\_128 model can achieve 28 milliseconds per inference with high accuracy (64.7\% on ImageNet validation dataset) on a single ARM CPU of Pixel 2. More details are described in the paper~\cite{ShengFeng2018MobileNets}.

\subsection{Third Prize of Track 1}

The Expasoft team wins the third prize of Track 1.  The team builds a neural network architecture that gives the high accuracy and inference time equal to 30 ms on Google Pixel~2. The team chooses two most promising architectures MobileNet and MobileNet-v2. Running MobileNet-v1\_224 with float-32 on Pixel-2 phone gives 70\% accuracy and inference time of 81.5 ms. The team chooses two main directions to accelerate the neural network:  quantization and reducing input image resolution. Both methods lead to accuracy reduction and the team finds trade-off for accuracy- speed relation.  Tests of MobileNet and MobileNet-v2 architectures suggest quantizing the neural networks into uint8 data format. The team's evaluation shows final score equal to 68.8\% accuracy and 29ms inference time. Table~\ref{table:track1thirdprize} compares the accuracy and time for different configurations.

Quantization to uint8 reduces inference time from 81ms to 68ms but leads to significant accuracy drops. During standard quantization process in Tensorflow, it is required to start from full-precision trained model and learn quantization parameters (min and max values). Instead of joint training of neural network and tuning quantization parameters, the Expasoft team proposes another approach:  tuning quantization parameters using Stochastic Gradient Descent approach with State Through Estimator~\cite{DBLP:journals/corr/BengioLC13}  of gradients of discrete functions (round, clip) without updating weights of the neural networks. Loss function for this process is L2 for embedding layers of full-precision and quantized networks. Mobilenet-v1 224 $\times$ 224 quantized such way shows 65.3\% accuracy. The proposed method requires more detailed research but has two significant advantages: (1) 
  The method doesn't require labeled data for tuning quantization parameters.
    (2) Training is faster  because there is no need to train neural network and only tuning is needed for quantizatizing parameters.

\subsection{First Prize of Track 2}

The Seoul National  University team wins the first prize in Track 2.  Optimization is needed to balance speed and accuracy for existing deep learning algorithms originally designed for fast servers. The team discovers that optimization efficiency differs from network to network. For this reason, the team compares various object detection networks on the Jetson TX2 board to find the best solution.
 
The team compares five one-stage object detectors: YOLOv2, YOLOv3, TinyYOLO, SSD, and RetinaNet. Two-stage detectors such as Faster R-CNN are excluded because they are slightly better in accuracy than the one-stage detectors but much slower in speed. The baseline for each network is selected as the authors' published form. For the comparison, the networks are improved with several software techniques briefly introduced as follows.
\begin{enumerate}
    \item Pipelining: An object detection network can be split into three stages: input image pre-processing, a convolutional network body, and post-processing with output results of the body. Since the network body is typically run on the GPU and the others are on the CPU, they can run concurrently with the  pipelining technique. 
    
    \item Tucker decomposition: It is one of the low-rank approximation techniques. As described in~\cite{Tucker1966}, 2-D Tucker decomposition is used in the network comparison.
    
    \item Quantization: For the Jetson TX2 device, only 16-bit quantization is allowed and it is applied to the networks. 
    
    \item Merging batch normalization layer into weights: Since the batch normalization layer is a linear transformation, it can be integrated into the previous convolutional layer by adjusting the filter weights before running the inference. 
    
    \item Input size reduction: This is an easy solution to enhance the network throughput. It was also observed in experiments that the effect of this technique depends on the networks.\\
\end{enumerate}

By comparing the networks, the team finds that \textit{YOLOv2} outperforms the other networks for the on-device object detection with the Jetson TX2.
Table~\ref{table:track2winner1} shows how much the YOLOv2 network is  enhanced by the series of the improvements. Since the total energy consumption is inversely proportional to the network speed, the score (mAP/Wh) can be estimated as mAP $\times$ speed. This optimized \textit{YOLOv2} network is selected for the LPIRC Track 2 competition.

\begin{table}[]
    \centering
    \begin{tabular}{|l|r|r|r|p{0.5in}|} \hline
         Description &
mAP(A) &
FPS(B) &
A x B & 
Normalized score \\ \hline

Baseline 
 &
51.1 &
7.97 &
407 & 
1.00 \\ 

Pipelining &
51.1 &
8.85 &
452 &
1.11 \\

Tucker &
50.2 &
15.1 &
758 &
1.86 \\

Quantization &
50.2 &
19.9 &
999 &
2.45 \\

256 x 256 &
43.0 &
32.5 &
1640 &
4.03 \\

Batch = 16 &
43.0 &
90.3 &
3880 &
9.54 \\ \hline

    \end{tabular}
    \vspace{0.1in}
    \caption{Different improvements and their scores by the Seoul National University team. The baseline is 416x416.
    FPS: frames per second.}
    \label{table:track2winner1}
\end{table}

YOLOv2 is tested on the Darknet framework in the experiments and it needs to be translated to Caffe2 framework for Track 2. The team implements custom Caffe2 operators to support Darknet-specific operators and optimization techniques such as pipelining and 16-bit quantization. Additionally, various batch sizes for the network are tested to find the best batch size for the submission.
Through the steps illustrated above, the estimated score for the YOLOv2 has increased about 9.54 times compared with the baseline and this result surpassed the other object detection networks on the Jetson TX2.

\subsection{Third Prize of Track 2}

The team's members are from Tsinghua University, University of Science and Technology of China, and Nanjing University. The team evaluates several mainstream object detection neural models and picks the most efficient one. The selected model is then fine-tuned with sufficient dataset before being quantized into 16-bit float datatype in order to achieve better power-efficiency and time-efficiency in the exchange of minimal accuracy loss.

The team explores popular object detection neural architectures such as YOLO, RCNN and their variants. Among these architectures, YOLO V3 achieves the best balance between accuracy and computational cost. However, considering the difficulty of engineering implementation in a short amount of time, the team chooses faster RCNN as the base structure and then quantizes the parameters in order to shorten the inference time and reduce power consumption.

Faster RCNN consists of 2 sub-networks: the feature extraction network and the detection network. While the latter doesn't seem to have much space for altering, there are many options for the former, such as VGG and MobileNet. The MobileNet family is known for their lower computational cost for achieving equivalent accuracy compared with traditional feature extraction architectures. Although MobileNets are reported with good classification results, the mAP for object detection seems low. The team decides to choose VGG-16.

\begin{figure}[h]
\centering
{\includegraphics[width=3.2in]{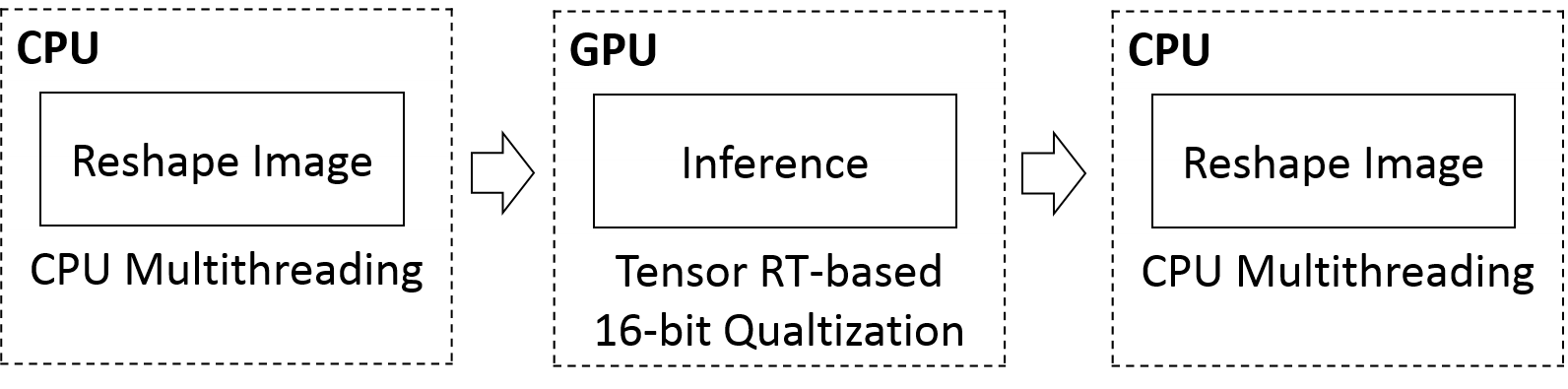}}
\caption{Pipeline used by the team of Tsinghua University, University of Science and Technology of China, and Nanjing University. }
\label{fig:track2third}
\end{figure}

The overview of the software optimization methods can be seen in Figure~\ref{fig:track2third}. The team reshapes the images on the CPU 
and then conducts the inference on the GPU. After the inference, the images are reshaped again to obtain the coordinates of the bounding boxes. The team applies three different techniques to accelerate the image recognition process: TensorRT-based inference, 16-bit Quantization, and CPU multithreading.

This implements the well-trained Fast-RCNN model used for image recognition on TensorRT. NVIDIA TensorRT 
is a high-performance deep learning inference optimizer and runtime  system that delivers low latency and high-throughput for deep learning inference applications. The inference time becomes much shorter by using TensorRT and therefore saves energy. TensorRT provides quantization operation for the GPU inference engine. The computation latency is shortened because of less floating point arithmetic. In order not to reduce the mAP of the model, the team quantizes the weight to 16-bit while doing inference.

The method reshapes the images before and after the inference process. Since reshaping on the CPU is costly, the method uses CPU multithreading to accelerate the process. The method generates two threads on each CPU core and each thread processes one batch of data. TX 2 has 6 CPU cores and the method creates 12 threads. The inference on the GPU can only work in a single thread; thus the method takes the inference as a mutual process and different threads need to compete for the GPU. Applying multithreading method results in a 40\%  speeding up (the fps without multithreading is 3.17 while the fps with multithreading is 4.41), not as high as expected.  The reason is that competition for the mutual recourse among the threads restricts acceleration. Different threads need to wait for the inference process because it needs mutual lock to avoid the hazard.

\subsection{First and Second Prizes of Track 3}

The ETRI and KPST team wins both the first and the second prizes of Track 3. The performance of object detection can be evaluated using accuracy, speed, and memory. The accuracy is a common measurement and it has been widely used in comparing different object detection architectures. However, its performance is dependent upon speed and memory as well as accuracy for the given environments and applications. In this competition, the performance is measured using accuracy (mAP) and energy consumption (WH). The accuracy is influenced by the detection architectures, an input resolution and a base network for feature extraction. The more complex structures with a high-resolution image may achieve higher accuracy at higher energy consumption. To obtain a high score, it is necessary to balance accuracy and energy consumption.  

The team examines the score function and finds  the most important factor is energy consumption. At the same accuracy, the score is higher at lower energy consumption. Moreover, the accuracy is weighted by the processed images ratio within 10 minutes. This means that the detection architecture should be fast and light and the trade-off between accuracy and energy consumption. Accordingly, single stage detectors such as SSD~\cite{10.1007/978-3-319-46448-0_2} and YOLO~\cite{8100173} are considered in the pipeline. The team selects SSD as the detection framework due to its simplicity and stable accuracy among feature extractors~\cite{8099834}. To obtain the best score, the team performs three optimization steps: (1)  detection structure optimization, (2) feature extraction optimization, (3) system and parameters optimization.

For detection structure optimization,  the original SSD is based on VGG and its performance is well balanced in accuracy and speed. Its speed and memory can be improved for low-power and real-time environments. To speed up, the team proposes efficient SSD (eSSD) by adding additional feature extraction layers and prediction layers in SSD. Table~\ref{table:track31st2nd} shows the comparison of SSD, SSDLite~\cite{Sandler2018CVPR}, and eSSD.  In SSD, additional feature extraction is computed by 1x1 conv and 3x3 conv with stride 2 and prediction uses 3x3 conv. The SSDLite replaces all 3x3 conv with depthwise conv and 1x1 conv. The eSSD extracts additional features with depthwise conv and 1x1 conv and predicts classification and bounding box of an object with 1x1 conv. This reduces memory and computational resources.

\begin{table}[]
    \centering
    \begin{tabular}{|l|p{1.2in}|p{1.2in}|} \hline
         
Type &
Additional 
 feature extraction &
Prediction \\ \hline

SSD & 
1x1 conv / 3x3 conv-s2 &
3x3 conv \\ \hline

SSDLite &
1x1 conv / 3x3 conv(dw)-s2 / 1x1 conv &
3x3 conv(dw) / 1x1 conv \\ \hline

eSSD &
3x3 conv(dw)-s2 / 1x1 conv &
1x1 conv \\ \hline

    \end{tabular}
    \vspace{0.1in}
    \caption{The comparison of feature extraction and prediction in SSD, SSDLite, and eSSD.}
    \label{table:track31st2nd}
\end{table}

Table~\ref{table:track31st2ndbase} shows accuracy, speed, and memory comparison of SSD variants in VOC0712 dataset. In this experiment, a simple prediction layer such as 3 by 3 or 1 by 1 is applied and an inference time (forward time) is measured in a single Titan XP (Pascal) GPU. SSDLite is more efficient in memory usage than eSSD, but eSSD shows better performance in speed than SSDLite.

\begin{table}[]
    \centering
    \begin{tabular}{|p{0.8in}|p{0.8in}|r|r|r|} \hline
         
Base 
Network
(300x300) &
Feature extraction / Prediction &
mAP &
Speed &
Memory \\ \hline

MobileNetV1 &
SSD /  3x3 &
68.6 &
8.05 &
34.8 \\ \hline
 
MobileNetV1 &
SSD / 1x1 &
67.8 &
6.19 &
23.1 \\ \hline

MobileNetV1 &
SSDLite /1x1 &
67.8 &
5.91 &
16.8 \\ \hline

MobileNetV1 &
eSSD / 1x1 &
67.9 &
5.61 &
17.4 \\ \hline

MobileNetV1 (C=0.75) &
eSSD / 1x1 &
65.8 &
5.20 &
11.1 \\ \hline

MobileNetV1 (C=0.75) &
eSSD(L=5) /1x1 &
65.8 &
4.62 &
10.9 \\ \hline

VGG &
SSD / 3x3 &
77.7 &
12.43 &
105.2 \\ \hline
    \end{tabular}
    \vspace{0.1in}
    \caption{The accuracy (mAP), speed (ms), and memory (MB) for different feature extraction and prediction architectures in VOC 0712 train and VOC 2007 test dataset.}
    \label{table:track31st2ndbase}
\end{table}

For feature extraction optimization, the base model of feature extractor, MobilNetV1~\cite{Howard2017arxiv} is used and feature extraction layers of MobileNetV1 is optimized in eSSD. To improve memory usage and computational complexity, the team uses 75\% weight filter channels (C=0.75). Although this drops accuracy, energy consumption is greatly reduced. The team uses five additional layers and modified anchors for a low resolution image. It generates a small number of candidate boxes and improves detection speed. After detection structures are optimized, the team also modifies MobileNetV1 by applying early down-sampling and weight filter channel reduction in earlier layers and trained the base model (MobileNetV1+) from scratch. All models are trained with ImageNet dataset and Caffe framework. In training, batch normalization is used and trained weights are merged into final model as introduced in~\cite{Fu2017arxiv}.

After feature extraction optimization, scores of proposed models are measured. Furthermore, the team compares these scores with scores of YOLOv3-tiny~\cite{redmon2018yolov3}. Comparison results are presented in Table~\ref{table:track3_yoloCompare}. Scores of proposed models are located in the range between 0.18 and 0.21. In contrast, scores of models from YOLOv3-tiny have a peak in the range of input image resolution between 256 and 224, and the score drops rapidly when the resolution decreases to 160. Scores of all three proposed models are higher than the best score from YOLOv3-tiny. In detail, comparing the proposed model of input resolution 224 and YOLOv3-tiny model of input resolution of 256, both have similar accuracy (25.7 vs 25.5) but the proposed model consumes less power about 17.5\%.

\begin{table}[]
    \centering
    \begin{tabular}{|p{1.2in}|r|r|r|r|} \hline
         
Models &
mAP &
sec &
Energy &
Score \\ \hline
eSSD-MobileNetV1+
(C=0.75, I=224,
th=0.01) &
25.8 &
476 &
1.244 &
0.2074 \\ \hline
eSSD-MobileNetV1+
(C=0.75, I=192,
th=0.01) &
23.3 &
445 &
1.170 &
0.1991 \\ \hline
eSSD-MobileNetV1+
(C=0.75, I=160, th=0.01) &
19.3 &
381 &
1.049 &
0.1840 \\ \hline

Yolov3-tiny 
(I=256, th=0.01) &
25.5 &
428 &
1.508 &
0.1691 \\ \hline
Yolov3-tiny 
(I=224, th=0.01) &
22.5 &
411 &
1.313 &
0.1714 \\ \hline  
Yolov3-tiny 
(I=160, th=0.01) &
15.7 &
330 &
1.082 &
0.1451 \\ \hline
         
    \end{tabular}
    \vspace{0.1in}
    \caption{Score of proposed models and YOLOv3-tiny. \textit{C, I, th} represent channel reduction, input resolution, Confidence threshold, respectively. The time is measured for processing 20,000 images.
}
    \label{table:track3_yoloCompare}
\end{table}

For system and parameters optimization, after training the models, the system is set up and the trained models are ported to NVIDIA TX2. In object detection, multiple duplicate results are obtained and Non-Maximal Suppression (NMS) with thresholding is important. The team tunes the NMS process between CPU and GPU to reduce computational complexity and adjust the confidence threshold to decrease result file size for the network bandwidth. Then batch size modification and queuing are applied to maximize speed in detection and to increase the efficiency of network bandwidth. After tuning the speed, to minimize energy consumption, the team uses the low power mode (max-Q mode) in NVIDIA Jetson TX2. 

Scores of the proposed models after system and parameters optimization are described in Table~\ref{table:track3_batchExp}. The team achieves 1.5x--2x higher score after optimization, but the order is reversed, i.e. after optimization, the model of input resolution 160 has the highest score which has the lowest score before optimization. Batch processing increases the throughput of the detection process. Accordingly, it improves the score. However, batching process with larger input resolution needs more memory access so it decreases the score. 

\begin{table}[]
    \centering
    \begin{tabular}{|p{1.2in}|r|r|r|r|} \hline
         
Models &
mAP &
sec &
Energy &
Score \\ \hline
eSSD-MobileNetV1+
(C=0.75, 224,
th=0.05, batch=64) &
24.7 &
448 &
0.780 &
0.3167 \\ \hline
eSSD-MobileNetV1+
(C=0.75, 192,
th=0.05, batch=64) &
22.2 &
415 &
0.617 &
0.3598 \\ \hline
eSSD-MobileNetV1+
(C=0.75, I=160, th=0.05, batch= 96) &
18.5 &
305 &
0.512 &
0.3613 \\ \hline

    \end{tabular}
    \vspace{0.1in}
    \caption{Score improvement after system and parameters optimization. Batch represents batch size.
}
    \label{table:track3_batchExp}
\end{table}

Table~\ref{table:track3final} shows final model specifications at on-site challenge. Scores of both models are further increased by 10--23\%. The result shows similar accuracy with in-house measurement described in Table~\ref{table:track3_batchExp}, but the power consumption is significantly decreased. The team hypothesizes the difference is caused by the  network condition and test dataset. 

\begin{table}[]
    \centering
    \begin{tabular}{|p{1.2in}|r|r|r|r|} \hline
         
Models &
mAP &
sec &
Energy &
Score \\ \hline
eSSD-MobileNetV1+
(C=0.75, I=160, th=0.05, batch= 96) &
18.318 &
300 &
0.4119 &
0.4446
(1st) \\ \hline
eSSD-MobileNetV1+
(C=0.75, 192,
th=0.05, batch=64) &
21.192 &
362 &
0.5338 &
0.3970
(2nd) \\ \hline

    \end{tabular}
    \vspace{0.1in}
    \caption{Proposed (eSSD-MobileNetV1+) Model specifications of on-site challenge .
}
    \label{table:track3final}
\end{table}

\subsection{Second Prize of Track 3}
The Seoul National  University team shares the second prize of Track 3 because the scores are very close. The team chooses the TX2 board as the hardware platform because of the GPU-based deep learning application. The object detection network is an improved \textit{YOLOv2} network. The \textit{C-GOOD framework}~\cite{Kang:2018:CCG:3240765.3240786} is used. Derived from Darknet, this framework helps explore a design space of deep learning algorithms and software optimization options. The team reduces the energy consumption by managing the operating frequencies of the processing elements. The team discovers the optimal combination of CPU and GPU frequencies.

\subsection{Low-Power Vision Solutions in Literature}

MobileNet~\cite{Sandler2018CVPR} and SqueezeNet~\cite{Iandola2016} are network micro-architectures that set the benchmark accuracy with low energy consumption. Without any optimization, MobileNet requires $81.5$ ms to classify a single image,  higher than the $30$ ms wall-time requirement for Track 1. However, speeding up the network by quantizing the parameters of MobileNet leads to large precision loss. In order to further the gains of the micro-architecture solution, the Track 1 winning team develop a method to use quantization on MobileNet that can perform inference in~$28$ ms, without much loss of accuracy. CondenseNet~\cite{Huang2017} and ShuffleNet~\cite{Zhang2018} are two neural network architectures that use group convolutions to reuse feature maps in order to improve parameter efficiency. These architectures can classify a single image in about $52.8$ ms. Faster inference and lower memory requirements are obtained with Binarized Neural Networks~\cite{XNOR}\cite{Cour2016}, where each parameter is represented with a single bit. The accuracy of such networks is lower, making them unsuitable for some applications. LightNN~\cite{Ding2017} and CompactNet~\cite{Goel2018} bridge the gap between Binarized Neural Networks and architectures like CondenseNet, by using a variable quantization scheme. 


Some studies design hardware for neural networks. For example, Lee et al.~\cite{8481682} design an accelerator with precision using different numbers of bits. The system can achieve 35.6, 14.5, and 7.6 frames per second for running VGG-16 when using 1, 4, and 8 bits.  It achieves 3.08 TOPS/W at 16-bit precision and 50.6 TOPS/W at 1-bit precision. Abtahi et al. ~\cite{8392465} compare specially-designed hardware for CNN with Jetson TX1 and Cortex A53, Zynq FPGA and demonstrate higher throughput and better energy efficiency. Gholami et al.~\cite{8575377} reduce the sizes of neural networks for mobile hardware with comparable or better accuracy with prior work. The smaller networks are faster and consume less energy. Even though these studies use ImageNet, they do not have the same settings as LPIRC. As a result, it is not possible making direct comparison with the winners' solutions.

Single stage object detection techniques, like SSD, Yolo~v2, Yolo 
~v3, and TinyYolo have an inherent advantage in terms of energy consumption over multiple stage techniques, like Faster RCNN. However, when using multiple stages it may be possible to obtain  significantly higher accuracy. The winners of Track~2 improve the single stage detector by using quantization and pipelining. Their technique outperforms all previous solutions on the score. Some other teams also use efficient network architectures in the two stage Faster~RCNN object detector to improve its execution speed and memory requirement.

\section{Industry Support for Low-Power Computer Vision}

\subsection{Google Tensorflow Models}

Google provided technical support to Track 1 in two aspects. First, Google open-sources TfLite, a mobile-centric inference engine for Tensorflow models. TfLite encapsulates implementations of standard operations such as convolutions and depthwise-convolutions that shield the contestants from such concerns. This  allows more people  to participate in the competition. Second, Google provides the mobile benchmarking system and it allows repeatable measurement of the performance metric. The system comes in two flavors: a stand-alone App that the contestants can run on their local phones, and a cloud-based service where the submissions will be automatically dispatched to a Pixel 2 XL phone and benchmarked using the standardized environment. The App 
and a validation tool 
are provided to facilitate rapid development, exploring novel models and catching runtime bugs. 
Once the model is sufficiently polished, it will be available as 
cloud service for refinements and verification.

\subsection{Facebook AI Performance Evaluation Platform}

Machine learning is a rapidly evolving area: new and existing framework enhancements, new hardware solutions, new software backends, and new models. With so many moving parts, it is  difficult to quickly evaluate the performance of a machine learning model. However, such evaluation is  important in guiding resource allocation in (1) development of the frameworks, (2) optimization of the software backends, (3) 
 selection of the hardware solutions, and (4) iteration of the machine learning models.
Because of this need, Facebook has developed an AI performance evaluation platform (FAI-PEP)
to provide a unified and standardized AI benchmarking methodology.
FAI-PEP supports Caffe2 and TFLite frameworks, the iOS, Android, Linux, and Windows operating systems. The FAI-PEP is modularly composed so that new frameworks and backends can be added easily. The built-in metrics collected by FAI-PEP are: latency, accuracy, power, and energy. It also supports reporting arbitrary metrics that the user desires to collect. With FAI-PEP, the benchmark runtime condition can be specified precisely, and the ML workload can be benchmarked repeatedly with low variance.

\section{Future of Low Power Computer Vision}

In 2018 CVPR, LPIRC invites three speakers from Google and Facebook sharing their experience on building energy-efficient computer vision. More than 100 people attended the workshop. The panel after the speeches answers many attendees' questions. The high participation suggests that there is strong interest, in both academia and industry, to create datasets and common platforms (both hardware and software) for benchmarking different solutions. Readers interested in contributing to future low-power computer vision are encouraged to contact the LPIRC organizers for further discussion.
Over four years, LPIRC has witnessed impressive improvement of the champions' scores by 24 times. The tasks in LPIRC--detecting object in images-- marks a major triumph in the vision technologies. However, many more challenges are still yet to be conquered. 

\subsection{Future Vision Challenges:  Action, Intention, Prediction, Emotion, Implication}

Visual data (image or video) has rich information that is often difficult to express by words. Here, the authors suggest possible topics for future competitions in low-power computer vision.  More specifically, we suggest research directions in computer vision beyond processing pixels: action, intention, prediction, emotion, and implication. 

\begin{figure}[h]
\centering
{\includegraphics[width=3in]{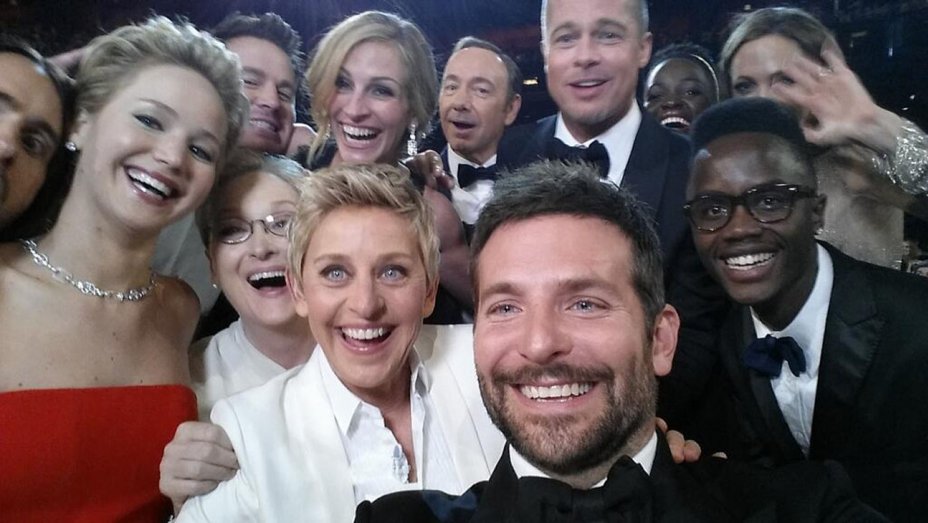}}
\caption{Selfie of Actors and Actresses after the 2014 Oscar Award Ceremony. Source: {\tt hollywoodreporter.com}}
\label{fig:oscarsselfie}
\end{figure}

Consider Figure~\ref{fig:oscarsselfie} as an example. This single image can provide rich information beyond detecting faces and recognizing names, even though both are important tasks in computer vision. The action here can be described as {\it laughing}, {\it gathering}, {\it sharing}.  What is their intention? Perhaps {\it celebrating}, {\it enjoying}.  This photograph was considered one of the most retweeted postings. Would it be possible to {\it predict} that this image would be popular becasue of the celebrities?  The emotion expressed in this photograph can be described as {\it happy}. Answering these questions would need to draw deeper knowledge beyond the pixels. Even if a computer vision program can recognize the faces and know these are actors and actresses, would it be capable of {\it predicting} that this image would be shared widely?

\begin{figure}[h]
\centering
\subfigure[]{\includegraphics[width=2.5in]{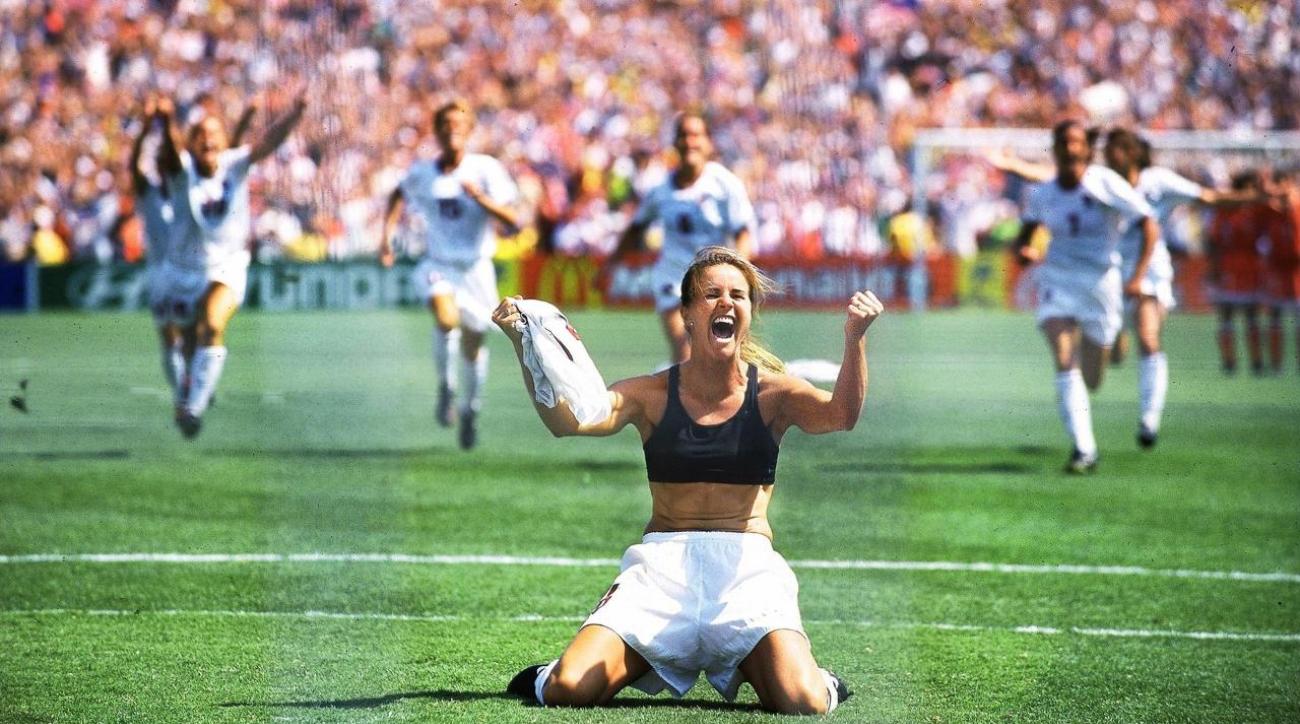}}
\subfigure[]{\includegraphics[width=0.5in]{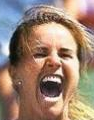}}
\caption{(a) Brandi Chastain celebrated after scoring in the penalty kick of the 1999 Women's World Cup Scooer. 
(b) The face of Chastain. Source: {\tt si.com}}
\label{fig:1999soccer}
\end{figure}

Next, consider the image shown in Figure~\ref{fig:1999soccer}. A vision task may be able to detect  people but a more interesting (and challenging) task is to describe the emotion: {\it happiness}, {\it joy},  {\it relieved}, {\it excited}, and {\it victorious}. This can be considerably difficult because the face of Chastain alone, shown in Figure~\ref{fig:1999soccer}~(b), may be insufficient to decide the emotion. Adding the  running team members at the background might be helpful  inferring the emotion. An even more difficult question is the {\it implication}: this team won and the other team lost. To obtain this information, it would be necessary to draw  the knowledge about the {\it context}: it is a soccer game. This may be possible from time and location when this image was taken. Many digital cameras (in particular smartphones) are equipped with GPS (global positioning systems) now and visual data can include the time and location. Such information can be used to look up other information, such as the event (e.g., a soccer game) and the participants (e.g., the teams). 

Creating a competition for evaluating computer vision's capability for understanding action, intention, prediction, emotion, or implication can be extremely difficult for three reasons.
First, the ``correct'' answers can be subjective or even dependent in culture, time, location, or other conditions. As a result, it would not be easy creating datasets for training or evaluation.
Second, even if the correct answers can be determined and agreed upon, labeling large amounts of data would require significant amounts of efforts.  Third, learning-based vision technologies need training examples; thus, vision technologies may be inadequate handling {\it exceptional} situations that rarely or have never occurred. In some cases (such as a natural disaster), data may have never been seen before. New thinking would be need to {\it expedite} training of unseen data for rescue operations.

\subsection{Low-Power Systems for Computer Vision}
The embedded computing systems for computer vision can be divided into three categories: 
(1)  \textit{General-purpose  systems:} CPU and GPU  are popular for computer vision tasks on embedded and mobile systems. Many low-power versions of CPU and GPU have been designed (e.g., NVIDIA's TX2) for this purpose. These designs often feature conventional low-power functions such as dynamic voltage and frequency scaling. Recently, they start to support functions in machine learning  such as low-precision arithmetics. (2)  \textit{Reconfigurable  systems:} FPGA is also a popular computing platform for  computer vision because it can be dynamically reconfigured to support different structures and precision~\cite{DBLP:journals/corr/abs-1809-00110}. FPGA has been deployed in data centers (e.g, Microsoft Azure) to support cloud tasks. Comprehensive software-hardware co-design flows have been also developed by leading FPGA companies to support machine learning  tasks. (3) \textit{Application-specific integrated circuit (ASIC):} ASIC offers the highest possible efficiency among all platforms but it needs the highest development cost and the longest design time. If the volume is large,  ASIC can become cost attractive. Huawei recently proposes Kirin 980 chipset to support AI tasks on their phones. Other examples include Google's Tensor Processing Unit (TPU) and Intel Nervana Neural Network Processor. Processors optimized for machine learning have been introduced by several vendors recently and these new systems  may appear in future LPIRC.

Among the above three technologies, GPU offers the most user friendly interface and comprehensive commercial support; FPGA offers the most flexible hardware reconfigurablity but requires complex hardware programming; ASIC offers the highest computing efficiency but also the longest development cycle. A major challenge in future low-power vision processing system  is the co-design between software and hardware, e.g., how to prune and quantify the network so that it can be efficiently mapped onto the hardware under certain constraints such as available memory capacity etc.

\section{Conclusion}

This paper explains the three tracks of the 2018 Low-Power Image Recognition Challenge. The winners describe the key improvements in their solutions. As computer vision is widely used in many battery-powered systems (such as drones and mobile phones), the need for low-power computer vision will become increasingly important. The initial success of the new tracks in 2018 also suggests the advantages of making focused advances on specific components of the vision system, as well as lowering the entry barrier to be inclusive of the general vision and machine learning communities. 

\section*{Acknowledgments}

IEEE Rebooting Computing is the founding sponsor of LPIRC. The sponsors since 2015 include Google, Facebook, Nvidia, Xilinx, Mediatek, IEEE Circuits and Systems Society, IEEE Council on Electronic Design Automation, IEEE GreenICT, IEEE Council on Superconductivity.  ETRI and KPST's work is supported by ETRI grant funded by the Korean government (2018-0-00198, Object information extraction and real-to-virtual mapping based AR technology). 

\bibliographystyle{unsrt}
\bibliography{ms}

\section*{Authors' Affiliations}

The authors are ordered alphabetically:
Sergei Alyamkin (Expasoft), 
Matthew Ardi (Purdue), 
Alexander C. Berg (University of North Carolina at Chapel Hill),  
Achille Brighton (Google), 
Bo Chen (Google), 
Yiran Chen (Duke),  
Hsin-Pai Cheng (Duke), 
Zichen Fan (Tsinghua University), 
Chen Feng (Qualcomm), 
Bo Fu (Purdue, Google), 
Kent Gauen (Purdue), 
Abhinav Goel (Purdue),
Alexander Goncharenko (Expasoft), 
Xuyang Guo (Tsinghua University), 
Soonhoi Ha (Seoul National University),
Andrew Howard (Google), 
Xiao Hu (Purdue),
Yuanjun Huang (University of Science and Technology of China), 
Donghyun Kang (Seoul National University), 
Jaeyoun Kim (Google), 
Jong Gook Ko (ETRI), 
Alexander Kondratyev (Expasoft), 
Junhyeok Lee (KPST), 
Seungjae Lee (ETRI), 
Suwoong Lee (ETRI), 
Zhiyu Liang (Qualcomm), 
Zichao Li (Nanjing University), 
Xin Liu (Duke), 
Juzheng Liu (Tsinghua University), 
Yang Lu (Facebook), 
Yung-Hsiang Lu (Purdue), 
Deeptanshu Malik (Purdue), 
Hong Hanh Nguyen (KPST), 
Eunbyung Park (University of North Carolina at Chapel Hill), 
Denis Repin (Expasoft), 
Tao Sheng (Qualcomm), 
Liang Shen (Qualcomm), 
Fei Sun (Facebook), 
David Svitov (Expasoft), 
George K Thiruvathukal (Loyola University Chicago and Argonne National Laboratory), 
Jingchi Zhang (Duke), 
Baiwu Zhang (Qualcomm), 
Xiaopeng Zhang (Qualcomm), 
Shaojie Zhuo (Qualcomm).

\end{document}